\documentclass[10pt,twocolumn,letterpaper]{article}

\usepackage{cvpr}
\makeatletter
\@namedef{ver@everyshi.sty}{}
\makeatother
\usepackage{tikz}              
\usepackage{times}
\usepackage{epsfig}
\usepackage{graphicx}
\usepackage{amsmath}
\usepackage{amssymb}
\usepackage[noline,boxed]{algorithm2e}
\usepackage[colorinlistoftodos]{todonotes}
\usepackage{multirow}

\usepackage[normalem]{ulem} 

\usepackage{pgfplots}
\pgfplotsset{compat=1.14} 

\usepackage{subfig}

\DeclareMathOperator*{\argmin}{\arg\!\min}

\newcommand\blfootnote[1]{%
  \begingroup
  \renewcommand\thefootnote{}\footnote{#1}%
  \addtocounter{footnote}{-1}%
  \endgroup
}

\usepackage[pagebackref=true,breaklinks=true,letterpaper=true,colorlinks,bookmarks=false]{hyperref}

\cvprfinalcopy 

\def\cvprPaperID{15} 

\ifcvprfinal\fi
\begin{document}

\title{Associative Embedding for Game-Agnostic Team Discrimination}

\author{Maxime Istasse$^*$, Julien Moreau$^*$, Christophe De Vleeschouwer\\
UCLouvain ICTEAM, Belgium\\
{\small $^*$These authors contributed equally to this work}\\
{\tt\small \{maxime.istasse,julien.moreau,christophe.devleeschouwer\}@uclouvain.be}
}

\maketitle

\begin{abstract}
    Assigning team labels to players in a sport game is not a trivial task when no prior is known about the visual appearance of each team. Our work builds on a Convolutional Neural Network (CNN) to learn a descriptor, namely a pixel-wise embedding vector, that is similar for pixels depicting players from the same team, and dissimilar when pixels correspond to distinct teams.
    The advantage of this idea is that no per-game learning is needed, allowing efficient team discrimination as soon as the game starts.
    In principle, the approach follows the associative embedding framework introduced in~\cite{Newell17a} to differentiate instances of objects. Our work is however different in that it derives the embeddings from a lightweight segmentation network and, more fundamentally, because it considers the assignment of the same embedding to unconnected pixels, as required by pixels of distinct players from the same team.
    Excellent results, both in terms of team labelling accuracy and generalization to new games/arenas, have been achieved on panoramic views of a large variety of basketball games involving players interactions and occlusions. This makes our method a good candidate to integrate team separation in many CNN-based sport analytics pipelines.
   
\end{abstract}

\vspace{-1em}
\blfootnote{This research is supported by the DeepSport project of the Walloon region, Belgium. C. De~Vleeschouwer is funded by the F.R.S.-FNRS (Belgium).}

\section{Introduction}
\label{sec:intro}

Team sports analytics has numerous applications, ranging from broadcast content enrichment to game statistical analysis for coaches~\cite{Chen16, Thomas17, Zheng16}. 
Assigning team labels to detected players is of particular interest when investigating the relationship between team positioning and sport action success/failure statistics~\cite{Bialkowski14, Hobbs18, Liu14},
but also for some specific tasks such as offside detection in soccer~\cite{Dorazio09}
or ball ownership prediction in basketball~\cite{Wei15}.

Many previous works have investigated computer vision methods to detect and track team sport players~\cite{Cioppa18,Dorazio09,Lu18,Lu13,Manafifard17,Parisot17,Tong11}.
They can detect individual players, but generally resort to unpractical manual intervention or to unreliable heuristics to adapt their processing pipeline to recognize the players' team. 
Specifically, they generally need human intervention to adjust the team discriminant features (\eg \mbox{RGB} histogram in~\cite{Lu13}, or \mbox{CNN} features in~\cite{Lu18}) to the game at hand~\cite{Bialkowski14, Liu14, Lu18, Lu13}. 
A few methods have attempted to derive game-specific team features in an automatic manner~\cite{Dorazio09, Tong11}.
They consider the unsupervised clustering of color histograms~\cite{Dorazio09} or bags of color features~\cite{Tong11} computed on the spatial support of the players that are detected in the game at hand. Those methods depend on how well color discriminates the two teams, but is also quite sensitive to  
occlusions and to the quality of player detection and segmentation~\cite{Manafifard17}. This probably explains why those previous works have been demonstrated in outdoor and highly contrasted scenes, as encountered in soccer for example. We show in Section~\ref{sec:eval} that those methods fail to address real-life indoor cases.

As observed in~\cite{Lu18}, indoor sports analytics have to deal with lower color contrast between players and background, and more dynamic scenes, with more frequent occlusions.
\cite{Parisot16, Parisot17} also point out the low illumination, the strong reflections induced by dynamic advertising boards, the severe shadows, the large player density and the lack of color discrimination in indoor scenes.

In our work, we do not arbitrarily select a handcrafted feature to discriminate the teams. We do not consider a framework that requires game-specific adjustment either.
Instead we adopt a generic learning-based strategy that aims at predicting a feature vector in each pixel, in such a way that, independently of the game at hand, similar vectors are predicted in pixels lying in players from a same team, while distinct vectors are assigned to pairs of pixels that correspond to distinct teams. In other words, we train a neural network to separate,
in an embedding space,
the pixels of different teams and to group those in common team.
A simple and efficient clustering algorithm can then be used to dissociate different teams in an image.
Hence, we do not rely on explicit recognition of specific teams, but rather learn how to map player pixels to a feature space that promotes team clustering, whatever the team appearance. Although teams change at each game, there is thus no need for fine tuning or specific manual annotation for new games.
The approach has been inspired by the associative embedding strategy recently introduced to discriminate instances in object detection problems~\cite{Newell17b, Newell17a}. However, differently from~\cite{Newell17b, Newell17a}, it is demonstrated using a lightweight \mbox{ICNet} convolutional neural network (opening broader deployment perspectives than the heavy stacked hourglass architecture promoted in~\cite{Newell17b, Newell17a}) and, to our knowledge, is the first work assigning similar embeddings to unconnected pixels, thereby extending the 
field of application of pixel-wise associative embedding.

To validate our method, we have trained our network on a representative set of images captured in a variety of games and arenas. Since only a few player keypoints (head, pelvis, and feet) have been annotated in addition to the player team index, the player segmentation component of our network has been trained with approximate ground-truth masks, corresponding to ellipses connecting the key points.
Our \mbox{CNN} model is validated on games (teams) and arenas that have not been seen during training. It achieves above $90\%$ team recognition accuracy, despite the challenging scenes (indoor, dynamic background, low contrast) and the inaccurate segmentation ground-truth considered during training. Interestingly, the lightweight backbone makes the solution realistic for real-time deployment.

Our paper is organized as follow. Section~\ref{sec:sota} reviews the related works associated to CNN-based sport analysis, segmentation, and associative embedding.
Section~\ref{sec:model} then introduces our proposed method, using a \mbox{ICNet} variant to both segment the players and compute pixel-wise team discriminant embeddings.
The experiments presented in Section~\ref{sec:eval} demonstrate the relevance of our approach, while conclusions and some perspectives are provided in Section~\ref{sec:conclu}.

\section{Related works}
\label{sec:sota}

Recent developments in computer vision make an extensive use of Convolutional Neural Networks~\cite{Russakovsky15}.
This section reviews the specific type of \mbox{CNNs}, named Fully Convolutional Network (\mbox{FCN}), that is used for image segmentation. It then introduces the recent associative embedding methods considered to turn object class segmentation into object instance segmentation.


\subsection{Fully Convolutional Network (\mbox{FCN})}

Fully Convolutional Networks are characterized by the fact that they output spatial feature maps, strictly computed by the recursive application of convolutional layers, generally completed with \mbox{ReLu} activation and batch-normalization or dropout regularization layers.

In recent works dealing with sport video analysis, \mbox{FCNs} have been considered for specific segmentation tasks, including player jersey number extraction~\cite{Gerke17}, soccer field lines and players segmentation~\cite{Cioppa18}. In~\cite{Lu18}, a two-steps architecture, inspired by~\cite{Yang16} and~\cite{Yu16}, is even proposed to extract players bounding-boxes with team labels. 
The network however needs to be trained on a game-per-game basis, which is impractical for large scale deployment.
None of these works is thus able to differentiate player teams without requiring a dedicated training for each game, as proposed in Section~\ref{sec:model}, where a real-time amenable \mbox{FCN} provides the player segmentation mask, as well as a pixel-wise team-discriminant feature vector.

There are two main categories of real-time \mbox{FCNs}: encoder-decoder networks and multi-scale networks.

Encoder-decoder architectures adopt the encoder structure of classification networks, but replace their dense classification layers by fully convolutional layers that upsample and convolve the coded features up to pixel-wise resolution.
\\
\mbox{SegNet} (Segmentation Network)~\cite{Badrinarayanan17} was the first segmentation architecture to reach near real-time inference.
It is a symmetrical encoder-decoder network, with skip connection of pooling indices from encoder layers to decoder layers.
\mbox{ENet} (Efficient Neural Network)~\cite{Paszke16} follows \mbox{SegNet}, but comes with various improvements, whose most prominant one is the use of a smaller decoder than the encoder.

Quite recently, several authors proposed to adopt multi-scale architectures to better balance accuracy and inference complexity. Considering multiple scales allows to exploit both a large receptive field and a fine image resolution, with a reduced number of network layers.
Among those networks, \mbox{ICNet} (Image Cascade Network)~\cite{Zhao18} is based on \mbox{PSPNet} (Pyramid Scene Parsing Network)~\cite{Zhao17}, a state-of-the-art network for non real-time segmentation. \mbox{ICNet} encodes the features at three scales. The coarsest branch is a \mbox{PSPNet}, while finer ones are lighter networks, allowing to infer segmentation in real-time.
Two-columns network~\cite{Wu17}, \mbox{BiSeNet} (Bilateral Segmentation Network)~\cite{Yu18}, \mbox{GUN} (Guided Upsampling Network)~\cite{Mazzini18} and \mbox{ContextNet}~\cite{Poudel18}
are composed of two branches.
\subsection{Associative embedding}

An embedding vector denotes a 
local descriptor that characterizes a signal locally in a way that can support a task of interest. Embeddings are thus not defined a priori. Instead, they are defined in an indirect manner, to support the task of interest.
In computer vision, \mbox{FCNs} have recently been considered to compute pixel-wise embeddings in a variety of contexts related to pixel clustering or pixel association tasks. In this context, \mbox{FCN} training is not supervised to output a specified value. Rather, \mbox{FCN} training supervises the relations between the embedded vectors, and checks that they are consistent with the task of interest.

In~\cite{Vondrick18}, the embedding vector is used to compute the similarity between two pixel neighborhoods from two distinct images, typically to support a tracking task. Interestingly, a proxy task that consists in predicting the (known) color of a target frame based on the color in a reference frame is used to supervise the training of the \mbox{FCN} computing the embeddings. Good embeddings indeed result in relevant pixel associations, and in accurate color predictions.   
This reveals that a \mbox{FCN} can be trained in an indirect manner to support various higher-level tasks based on richer pixel-wise embedding.

Of special interest with respect to our team discrimination problem, \emph{associative} embeddings have been introduced in~\cite{Newell17b, Newell17a} and used in~\cite{Law18, Newell17b, Newell17a} to associate pixels sharing a common semantic property, namely the fact that they belong to the same object instance. Authors in~\cite{Newell17a} introduced associative embedding in the context of multi-person pose estimation from joints detection and grouping, and extended it to instance segmentation.
More recently, \cite{Law18} proposed \mbox{CornerNet}, a new state-of-the art one-shot bounding box object detector, by using associative embedding to group top-left and bottom-right box corners.
In all these publications, the network is trained to give close embeddings to pixels from the same instance and distant embeddings to pixels corresponding to different instances. All these works are based on the same heavy stacked hourglass architecture.
However, \cite{Newell17b} suggest that the approach is not strictly restricted to this architecture, as long as two important properties are fulfilled: first, the network should have access both to global and local information; second, pixel-wise prediction at fine resolution is recommended, in order to avoid that a vector is subject to concurrent instances. This makes \mbox{ICNet} a premium candidate to segment players and compute team-specific embeddings in real time, since it computes features at three scales instead of two for other lightweight multi-branch \mbox{FCN} architectures.

\section{Team segmentation using pixel-wise associative embedding}
\label{sec:model}

Player team discrimination is not a conventional segmentation problem since the visual specificities of each class are not known in advance. This section explains how associative embedding can be combined with player segmentation to address this problem.

\begin{figure*}
    \centering
    \includegraphics[width=0.9\linewidth,page=1]{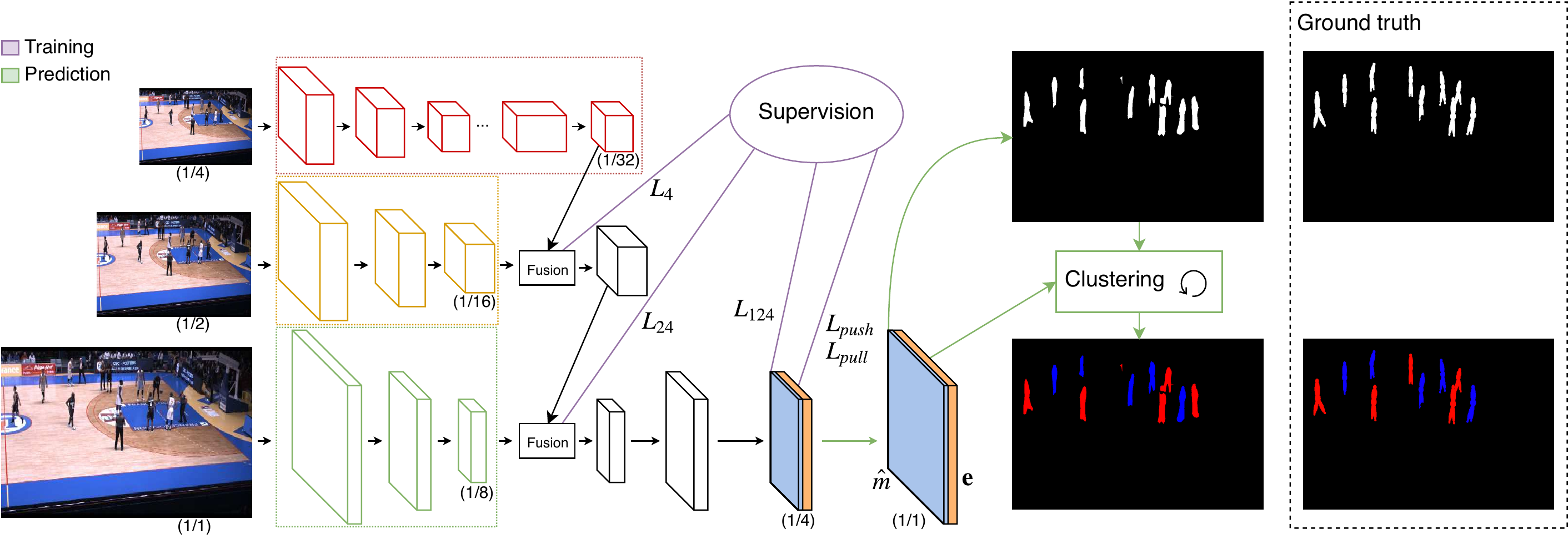}
    \caption{Overview of our architecture. \mbox{ICNet}~\cite{Zhao18} is used as backbone for following assets: pixel-wise segmentation, combination of three scales to encode global and local features, fast (\cite{Zhao18}~reaches 30~FPS at $1024 \times 2048$ resolution). Its last convolution is modified to output a segmentation mask along with vector embeddings in each pixel. We keep the multi-scale supervision for the segmentation and add $L_{push}$ and $L_{pull}$ to obtain similar embeddings in pixels of the same team and distant embeddings for pixels of different teams.
    After network inference, a simple clustering algorithm can effectively split in different teams the pixels from the segmentation mask $\hat{m}$ according to their embeddings $\mathbf{e}$.}
    \label{fig:network}
\end{figure*}

\subsection{Team discrimination \& player segmentation}

We propose to adopt the associative embedding paradigm to support the team discrimination task. In short, we design a fully convolutional network so that, in addition to a player segmentation mask, it outputs for each pixel a \mbox{$D$-dimensional} feature vector that is similar for pixels that correspond to players of the same team, while being distinct for pixels associated to distinct teams. As explained in the previous section, embeddings learning is not based on an explicit supervision. Instead, embeddings are envisioned as a latent pixel-wise representation, trained to support a pixel-wise association task, typically to group~\cite{Law18} or match~\cite{Vondrick18} pixels together. 
In the context of object detection, associative embedding has been applied with success in~\cite{Law18, Newell17a} to group pixels corresponding to a same object instance. In these works, multiple hourglass-shaped networks are stacked recursively in order to progressively refine the \mbox{1-D} embedding value that aims to differentiate object instances in a given class. Our work differs from~\cite{Newell17a, Newell17b} and~\cite{Law18} in two main aspects.

First, and because we target real-time deployment, the stacked hourglass architecture is replaced by an \mbox{ICNet}~\cite{Zhao18} backbone, as illustrated in Figure~\ref{fig:network}.
As stated in \cite{Zhao18}, \mbox{ICNet} reaches 30~FPS for images of $1024 \times 2048$ pixels on one Titan~X GPU card.
We use \mbox{ICNet} because its multi-scale encoders, along with a spatial pyramidal pooling, give access to a reasonably large receptive field (important to share embedding information spatially) while preserving the opportunity to exploit high-resolution image signal locally (important for a fine characterization of the content).

Second, our work deals with the problem of associating pixels of players that are scattered across the whole image. This is in contrast with the association  of neighboring/connected pixels generally considered in traditional association tasks~\cite{Law18, Newell17a}. 

\subsection{Network architecture}

The \mbox{ICNet} network architecture has mostly been left unchanged. Only the final convolution layer has been adapted to provide $D+1$ channels. Those comprise $1$ channel for semantic segmentation, with a sigmoid activation, along with $D$ channels for embeddings with linear activation. 
Figure~\ref{fig:network} presents the player segmentation channel in blue while the $D$ channels for embeddings are represented in orange.
A number of loss functions are combined to train the network.
Along with the multi-scale semantic segmentation loss from~\cite{Zhao18}, composed by $L_{124}$, $L_{24}$ and $L_4$, we add an embedding loss inspired by~\cite{Newell17a, Newell17b, Law18}. It comprises two components, $L_{pull}$ and $L_{push}$, which respectively pull teammates embeddings together and push opponents embeddings away from each other.
$L_{pull}$ and $L_{push}$ only apply to the finest resolution.
We have defined all loss components based on mean square distances.
\\
$L_{124}$, $L_{24}$ and $L_4$ losses are defined as:
\begin{equation}
    L_{s \in \{124,24,4\}} = \frac{1}{HW}\sum_{(i,j)}^{H\times W} (\hat{m}^s_{ij} - m^s_{ij})^2
\end{equation}
with $H$ and $W$ being the layer height and width,
while $\hat{m}^s$ and $m^s$ respectively denote the predicted and ground-truth player masks at scale $s$.
Similarly, $L_{pull}$ is formulated as:
\begin{equation}
    L_{pull} = \frac{1}{HW}\sum_{n}^{\{1,2\}} \sum_{(i,j)}^{M_n} (t_{ij}-T_n)^2,
\end{equation}
where $T_n$ is the mean of the embeddings $t_{ij}$ predicted across the pixels $M_n$ of team $n$, \ie $T_n = \sum_{(i,j)\in M_n} t_{ij}$.
\\
In~\cite{Newell17a}, the push loss is expressed as a mean over pairs of pixels of a cost function that is chosen to be high (low) when pixels that are not supposed to receive the same embedding have a similar (different) embedding.
Recently, \cite{Newell17b} and~\cite{Law18} employed a "margin-based penalty", and wrote that this is the most reliable formulation they tested.
Hence, we also adopt a margin-based penalty loss.
Formally, $L_{push}$ is defined similarly to $L_{pull}$, except that rather than penalizing embeddings that are far away from their centroid, it penalizes embeddings that are too close from the centroid of another team:
\begin{equation}
    L_{push} = \frac{1}{HW}\sum_{n}^{\{1,2\}}\sum_{(i,j)}^{M_n} \max(0;1-(t_{ij}-T_{3-n})^2)
\end{equation}
\\
Our global objective function finally becomes:
\begin{equation}
    \begin{split}
        L = &\lambda_{124} L_{124} + \lambda_{24} L_{24} + \lambda_{4} L_{4}\\
          + &\lambda_{pull} L_{pull} + \lambda_{push} L_{push}
    \end{split}
    \label{eq:loss}
\end{equation}
with the lambda loss factors having to be tuned (chosen values are explained in Section~\ref{ssec:implem}).

At inference, upsampling of last layer is inserted before activation (respectively bilinear and nearest neighbor interpolations for segmentation and embedding channels).
Then, a clustering algorithm is required to group pixels in teams.
Fortunately, as observed in~\cite{Newell17a}, the network does a great job at separating the embeddings for distinct teams, so that a simple and greedy method such as the one detailed in Algorithm~\ref{alg:clustering} is able to handle the clustering properly. As appears from the pseudocode, our naive clustering algorithm relies on the assumption that
a player pixel embedding surrounded by similar embeddings is representative of its team embedding.
Given a team embedding vector, player pixels are likely to be assigned to that team if their embedding lies in a sphere of radius $1$ around the team embedding.
We incorporate a refinement step in which we compute the centroid of the selected pixels.
Then, to resolve ambiguities, player pixels are associated to the closest of the centroids.


\begin{algorithm}
    \DontPrintSemicolon
    \newcommand{\emb}{\mathbf{e}}
    \newcommand{\cen}{\mathbf{c}_n}
    \SetKwInOut{AData}{Input}
    \SetKwInOut{AResult}{Result}
    \SetKwInOut{AInit}{Init}
    \AData{$\hat{m}_{ij}$ the predicted segmentation mask \\ $\emb_{ij}$ the predicted embedding in pixel (i,j)}
    \AResult{$O \in [\![ 0; 2 ]\!]^{H\times W}$ teams occupancy map ($0$ is associated to the background)}
    \BlankLine
    $\mathcal{N}(i,j)$ neighborhood of pixel $(i,j)$ \;
    $V_{ij} \gets \frac{1}{\lvert\mathcal{N}(i,j)\rvert} \sum_{(k,l)\in\mathcal{N}(i,j)} \lVert\emb_{ij} - \emb_{kl}\rVert_2^2$ \;
    $n \gets 0$ the team counter \;
    $\mathcal{R} \gets \mathcal{S} \gets \{(i,j) \mid \hat{m}_{ij} > 0.5\}$ the pixels to cluster \;
    $O \gets D_{1} \gets D_{2} \gets \mathbf{0}^{H\times W}$ \;
    
    \While{$\mathcal{R} \neq \{\}$ and $n<2$}{
        $n \gets n+1$ \;
        $(i, j) \gets \argmin_{(i,j) \in \mathcal{R}} V_{ij}$ \;
        $\cen \gets \emb_{ij}$ the centroid for team n \; 
        $\mathcal{M}_n \gets \{(i,j) \in \mathcal{R} \mid \lVert \emb_{ij}-\cen\rVert_2^2 < 1\}$ \;
        \BlankLine
        
        $\cen \gets \frac{1}{|\mathcal{M}_n|}\sum_{(i,j)\in \mathcal{M}_n} \emb_{ij}$ \;
        $\mathcal{M}_n \gets \{(i,j) \in \mathcal{R} \mid \lVert \emb_{ij}-\cen\rVert_2^2 < 1\}$ \;
        \BlankLine
        
        $\mathcal{R} \gets \mathcal{R} \setminus \mathcal{M}_n$ \;
        
        \For{$(i,j) \in \mathcal{S}$}{
            $ D_{n}(i,j) \gets \lVert \emb_{ij}-\cen\rVert_2^2 $ \;
        }
    }
    \For{$(i,j) \in \mathcal{S}$}{
        \uIf{$n = 1$}{
            $ O(i,j) \gets 1 $ \;
        }
        \ElseIf{$n = 2$}{
            $ O(i,j) \gets \argmin( \{ D_1(i,j); D_2(i,j) \} ) +1 $ \;
        }
    }
    \caption{Simple clustering algorithm of pixels in the space of their associated embeddings.
    Up to two centroids are searched and refined, from the observation that for a team, embeddings of neighbour pixels can serve as initial prototype when they are similar.
    Embeddings similarity in the neighborhood of pixel $(i,j)$, $\mathcal{N}(i,j)$, is called $V_{ij}$.
    After that, points are clustered according to their embedding vector's distance to centroids mapped in $D_{n}$ arrays.}
    \label{alg:clustering}
\end{algorithm}

\subsection{Implementation details and hyperparameters}
\label{ssec:implem}

Our network is trained to extract players only, and to estimate associative embeddings for team discrimination.
Referees and other non-player persons are part of the background class.
Our work is based on the \mbox{PyTorch} \mbox{ICNet} implementation~\cite{mshahsemseg}.
Parameters have been empirically tuned.
For the training, we employ Adam optimizer~\cite{Kingma15}.
Losses factors defined in Equation \ref{eq:loss} are:
$\lambda_{124} = 1$ and $\lambda_{24} = \lambda_{4} = 0.4$ as in original \mbox{ICNet}~\cite{Zhao18},
$\lambda_{pull} = \lambda_{push} = 4$ and are thus very different than in~\cite{Newell17a, Newell17b, Zhao18} because our pull and push losses definitions are averaged over pixels rather than over instances.
Our best found learning rate is $lr = 10^{-3}$, and has been implemented with the "poly" learning rate decay taken from~\cite{Zhao17, Zhao18, Yu18} and their own sources. 
Compared to them, we apply the decay by epochs instead of iterations, but we keep the same power of 0.9. Hence, the learning rate at $k^{\text{th}}$ epoch is
$lr . (1 - \frac{k}{max})^{power}$, with $max=200$ denoting the total number of epochs, and $lr$ being the base learning rate defined above.
All but last layers of \mbox{ICNet} are initialized with pretrained Cityscapes (\cite{Cordts16}) weights from~\cite{Zhao18}, but a full training is done as the point of view adopted for sport field coverage is too different from the frontal point of view considered by cars in Cityscapes.
Minibatch size is 16 and batch-normalization is applied.
Neither weight decay regularization, nor dropout are added, but the following random data augmentation is considered:
mirror flipping, central rotation of maximum 10 degrees, scaling such that $\min(\text{width}, \text{height}) = \text{random}([\frac{2}{3},\frac{3}{2}]) \times 512$, color jitter in the perceptually uniform \mbox{CIE~L*C*h} 
color space fixed to L~$\pm 10$, C~$\pm 7$ and h~$\pm 30$ degrees, to keep natural colors.
We trained the network on crops of $512 \times 512$ pixels, located randomly in scaled images.
Validation is performed on $512 \times 512$ pixels patches, extracted from images scaled such as its  $\min(\text{width}, \text{height})$ equals 512.
For each model, we select the parameters of the best epoch according to a validation score defined as the mean of intersection over union of the two teams, between prediction and our approximate reference masks.
Inference for testing is done on court images downsampled to $1024 \times 512$ and padded to preserve the aspect ratio.

In our implementation, we adopted \mbox{$5$-D} embeddings, mainly because more dimensions a priori get more ability to capture/encode visual team characteristics unambiguously.
We expect this ability to become especially useful when the receptive field does not cover the whole scene. In that case, the embedding prediction in one pixel may not be able to rely on a teammate appearance or on the absence of collision with an opponent embedding when those players are far and disconnected from the pixel of interest. The embeddings have thus to be consistent across the scene, despite their relatively local receptive field. In other words, they have to capture local team characteristics unambiguously.
In practice, \mbox{ICNet} builds a global receptive field, and our trials provided similar results with 1- to \mbox{5-D} embeddings.

\section{Experimental validation}
\label{sec:eval}

To assess our method, this section first introduces an original dataset, and associated evaluation metrics. It then runs a \mbox{K-fold} cross-validation procedure, and compares the performance of our associative embedding 
team discrimination, with a conventional color histogram clustering, 
applied on top of instance segmentation.

\subsection{Dataset characteristics}
\label{ssec:feet}

To demonstrate our solution, we have considered a proprietary basketball dataset. 
It involves a large variety of games and sport halls: images come from 40 different games and 27 different arenas. 
Images show innumerable situations: occlusions between teammates and/or opponents, regular player distribution, absence or presence of all the players, images from training sessions and professional games with public, various game actions, still and moving players, presence of referees, managers, mascots, dynamic led advertisements, photographers or other humans, various lighting conditions, different image sizes (smaller dimension is generally close or superior to 1000 pixels).
This dataset is composed of 648 images covering a bit more than half of the sport field.
Each player has been manually annotated. Annotations considered in our work include a team label (Team~A vs. Team~B), and an approximate player mask.
This mask has been derived from manual annotation of head, pelvis, and feet.
It consists in seven ellipses approximately covering the head, the body (between head and pelvis), the pelvis, the legs (between pelvis and each foot), and the feet.
Occlusions between ellipses of players located at different depth has been taken into account.
Similarly to~\cite{Cioppa18}, our experiments reveal that the network can learn despite the coarseness of the masks.
Players size in images feeding the network (scaling strategy in Section~\ref{ssec:implem})
is around $25 \times 75 \pm 15$ pixels.

\subsection{Evaluation metrics}

Our network enables player segmentation, as well as team discrimination.
Evaluation metrics should thus reflect whether players have been properly detected, and whether teammates have received the same team label. 
Therefore, we consider the following counters and metrics, to be computed on a set of test images:
\begin{itemize}
    \item $N_{miss}$: Number of missing players
    \item $N_{corr}$: Number of correct team associations
    \item $N_{err}$: Number of incorrect team associations
    \item Missed players rate,
    \begin{equation}
        R_{miss} = \frac{N_{miss}}{N_{corr}+N_{err}+N_{miss}}
    \end{equation}
    \item Correct team assignments rate,
    \begin{equation}
        R_{CTA} = \frac{N_{corr}}{N_{corr}+N_{err}}
    \end{equation}
\end{itemize}


We now explain how the outputs of our network, namely the player segmentation mask and the map of team labels derived from the embeddings clusters, are turned into those evaluation metrics\footnote{Since accurate ground truth segmentation masks are not available from the dataset (see Section~\ref{ssec:feet}),
the segmentation quality can not be assessed based on conventional intersection over union metrics.}. Given a reference segmentation mask and a team label for each player instance, a simple majority vote strategy is adopted. A player is considered to be detected when the majority of pixels in the player instance segmentation mask are part of the segmentation mask predicted by the network. In that case, the majority label observed in the instance mask defines the team of the player.
In practice, since our ground-truth mask only provides a rough approximation of the actual player instance silhouette, we resort to the part of the instance mask that is the most relevant for team classification, \ie to the two ellipses that respectively cover the body and the pelvis area. Since pixels that are in the central part of the body and pelvis ellipses are less likely to be part of the background, only the pixels that are sufficiently close to the main principal axis of the body/pelvis shape are considered.
(A distance threshold equal to one third of the maximal distance between ellipse border and principal axis has been adopted. Changing this threshold does not impact significantly the results.)


\subsection{Results}
\label{ssec:results}



In order to validate the proposed team discrimination method with available data,
we consider a \mbox{K-fold} cross-validation framework.
It partitions the \mbox{648-images} dataset into K disjoint subsets, named folds. Each \mbox{K-fold} iteration preserves one fold for the test, and use the other folds for training and validation. Average and standard deviation metrics can then be computed based on the K iterations of the training/testing procedure. In our case, ten folds of approximately equal size have been considered. Moreover, to assess whether the model generalizes properly on new games and new arenas, we construct the folds so that each fold contains images from distinct games and/or arenas. 
Table~\ref{tab:game-kfold} lists cross-game folds characteristics,
and Table~\ref{tab:hall-kfold} cross-arena folds characteristics.
%

To estimate the value to give to our results, we compare them to a baseline reference. Since most previous methods recognize teams based on color histograms~\cite{Dorazio09,Lu13,Tong11}, generally after team-specific training, we compare associative embeddings to a method that collects color histograms on player instances, before clustering them into two sets. In practice, as for the associative embedding evaluation, only the player pixels that are sufficiently close to the body/pelvis principal axis are considered to build the histogram in \mbox{RGB}, with 8 bins per dimension (\mbox{512-dimensional} histogram).
Adopted clustering is the~\cite{scikit-learn} implementation of variational inference algorithm with a Dirichlet process prior~\cite{Blei06}, to fit at max two gaussians representing our two clusters (two teams). This method has the advantage of being able to automatically reduce the number of prototypes, it is useful when less than two teams are visible in an image.

\begin{table}
    \centering
    \begin{tabular}{|l|cccc|}
        \hline
         Fold & 1 .. 3 & 4 & 5 .. 9 & 10 \\
         \hline
         \hline
         Train & 516 & 518 & 520 & 518 \\
         Val & 66 & 64 & 64 & 66 \\
         Test & 66 & 66 & 64 & 64 \\
         \hline
    \end{tabular}
    \caption{Splits of the dataset used for cross-game validation. Each column corresponds to one (set of) folds, and lines define the number of training/validation/test samples.
    Validation and test sets contain the images from 4 games.
    }
    \label{tab:game-kfold}
\end{table}
\begin{table}
    \centering
    \begin{tabular}{|l|cccccc|}
        \hline
         Fold & 1 & 2 .. 5 & 6 & 7 & 8 .. 9 & 10 \\
         \hline
         \hline
         Train & 514 & 516 & 518 & 522 & 524 & 518 \\
         Val & 66 & 66 & 64 & 62 & 62 & 68 \\
         Test & 68 & 66 & 66 & 64 & 62 & 62 \\
         \hline
    \end{tabular}
    \caption{Splits of the dataset used for cross-arena validation. Each column corresponds to one (set of) folds, and lines define the number of training/validation/test samples.
    Validation and test sets contain the images from 2 or 3 halls.
    } 
    \label{tab:hall-kfold}
    \vspace{-0.3em}
\end{table}


Results of cross-game validation are presented in Table~\ref{tab:game-eval},
while cross-validation on sport halls is presented in Table~\ref{tab:hall-eval}.
Standard deviations are low, demonstrating the weak dependence to a specific set of training data.
Rate of missing detections is about 11\%, which is an acceptable rate considering our backbone is the real-time \mbox{ICNet} model~\cite{Zhao18} with arduous indoor basketball images.
It could probably be improved with a finer tuning of hyperparameters, as well as more accurate segmentation masks and a formulation that involves a class for referees (see failure cases analysis below). More recent and effective improved segmentation networks could also be considered as long as they are compatible with associative embedding.

\begin{figure*}
    \centering
    \vspace{-1em}
    \subfloat{ \includegraphics[width=0.3\linewidth]{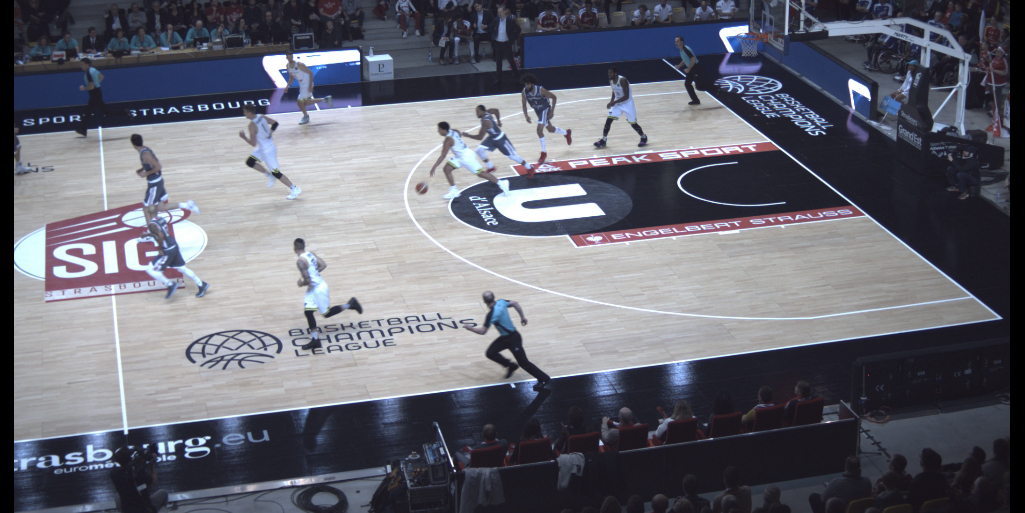} }
    \subfloat{ \includegraphics[width=0.3\linewidth,
                trim={20mm 45mm 100mm 15mm},clip]{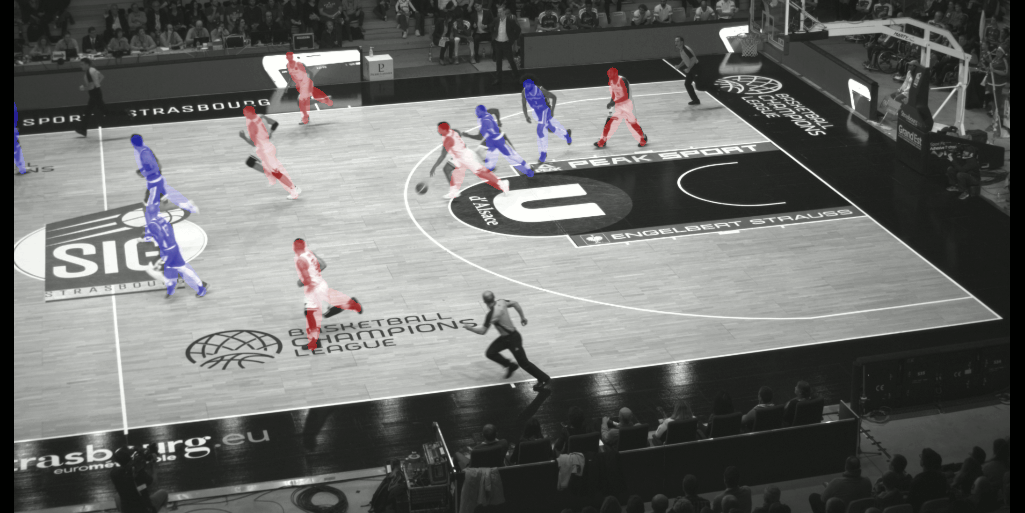} }
    \subfloat{ \includegraphics[width=0.3\linewidth,
                trim={20mm 45mm 100mm 15mm},clip]{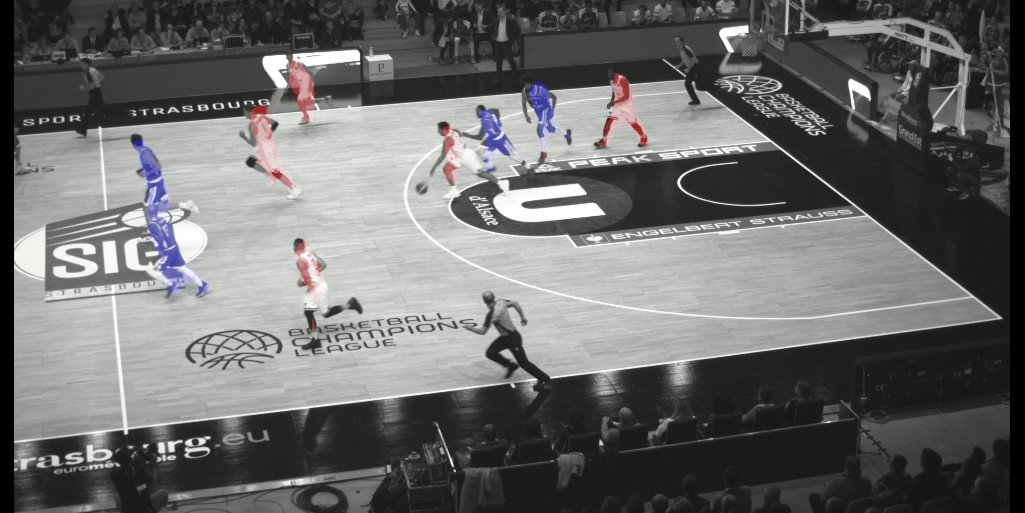} }
    \\
    \vspace{-0.3em}
    \subfloat{ \includegraphics[width=0.3\linewidth]{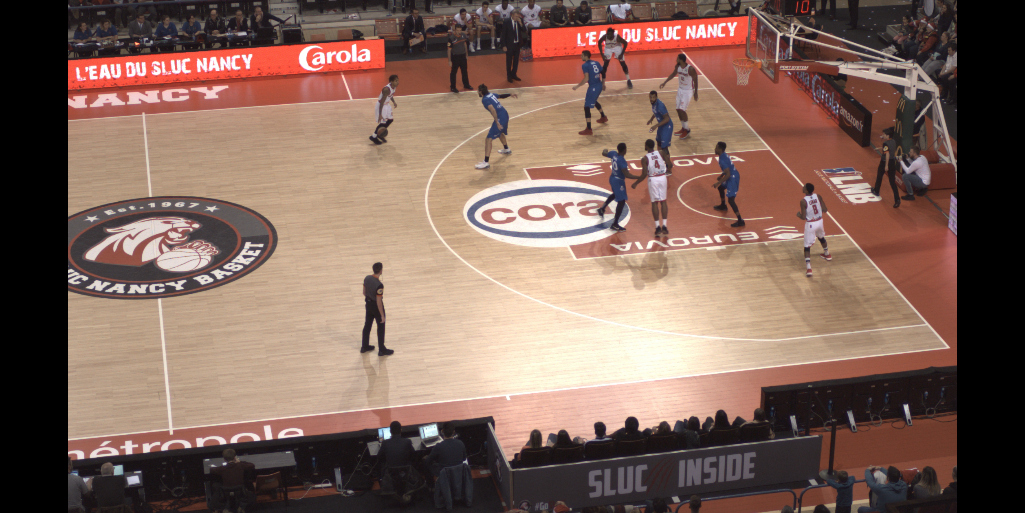} }
    \subfloat{ \includegraphics[width=0.3\linewidth,
                trim={120mm 70mm 40mm 10mm},clip]{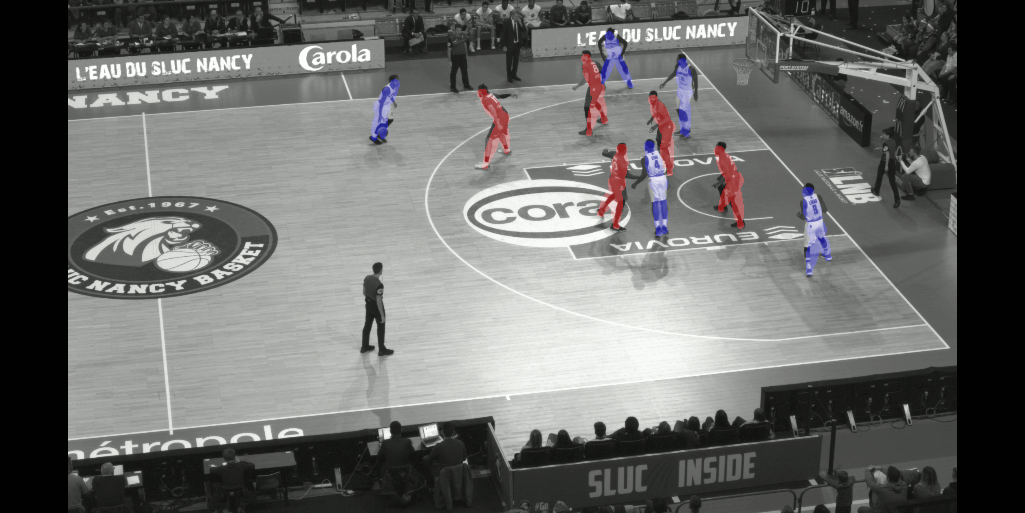} }
    \subfloat{ \includegraphics[width=0.3\linewidth,
                trim={120mm 70mm 40mm 10mm},clip]{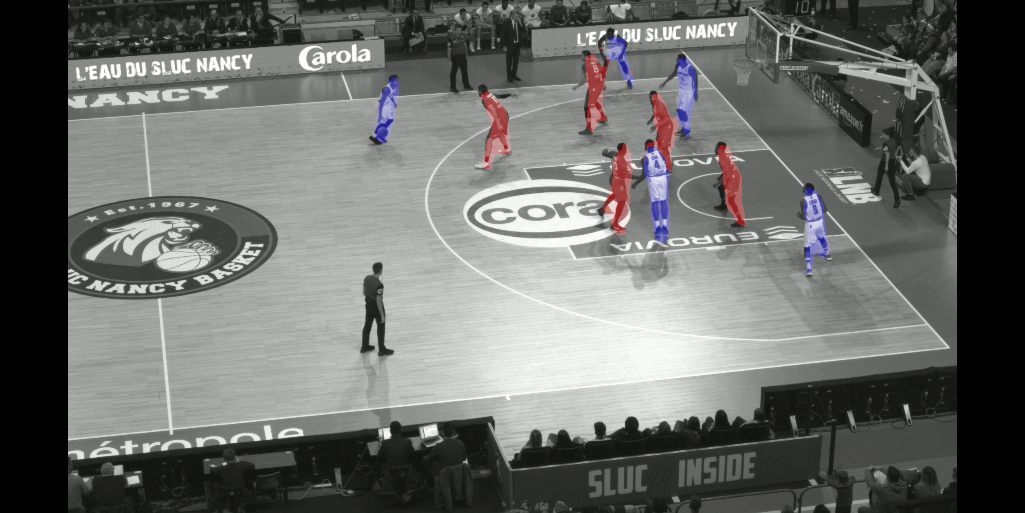} }
    \\
    \vspace{-0.3em}
    \subfloat{ \includegraphics[width=0.3\linewidth]{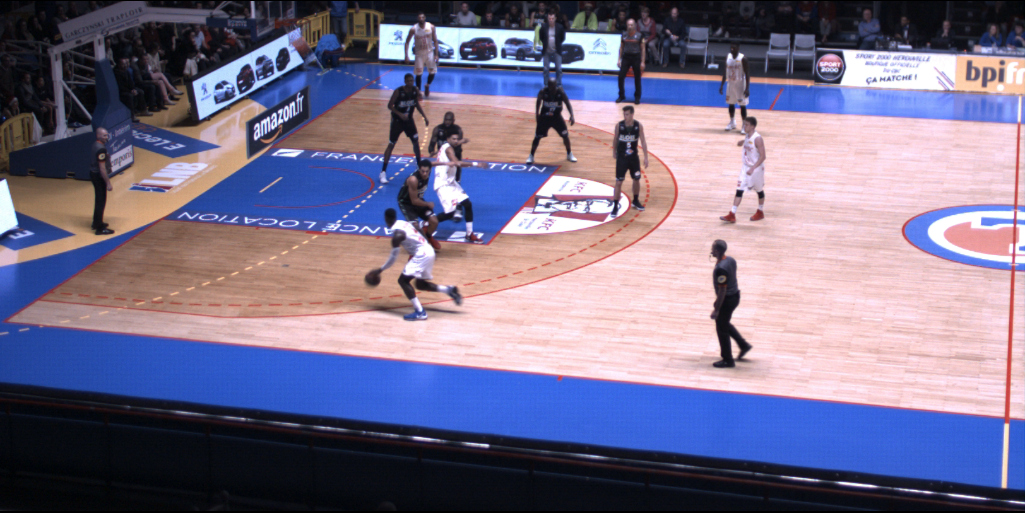} }
    \subfloat{ \includegraphics[width=0.3\linewidth,
                trim={25mm 50mm 75mm 0mm},clip]{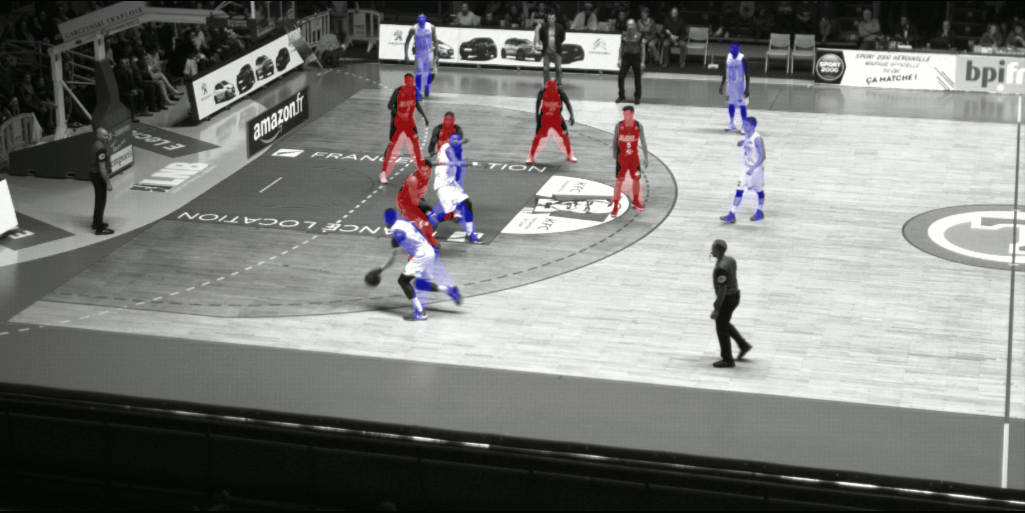} }
    \subfloat{ \includegraphics[width=0.3\linewidth,
                trim={25mm 50mm 75mm 0mm},clip]{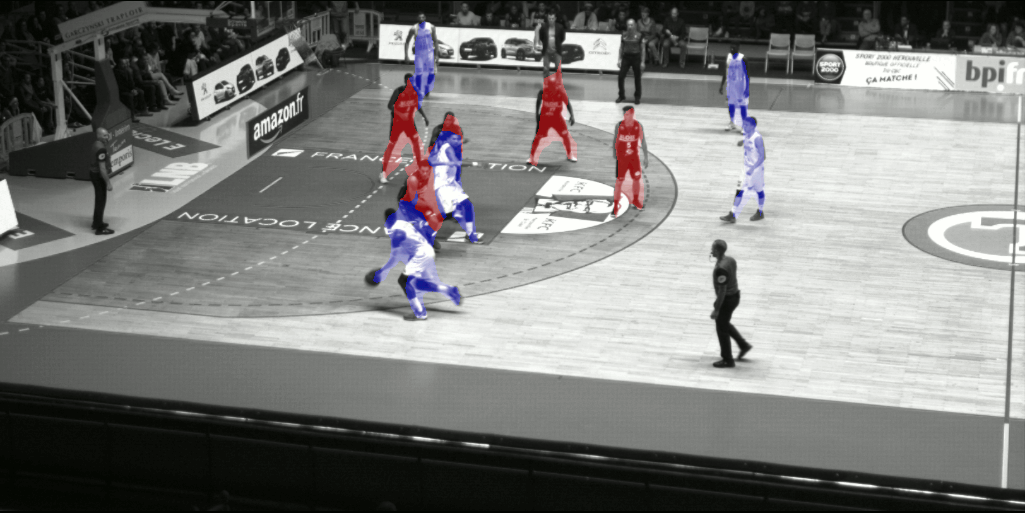} }
    \\
    \vspace{-0.3em}
    \subfloat{ \includegraphics[width=0.3\linewidth]{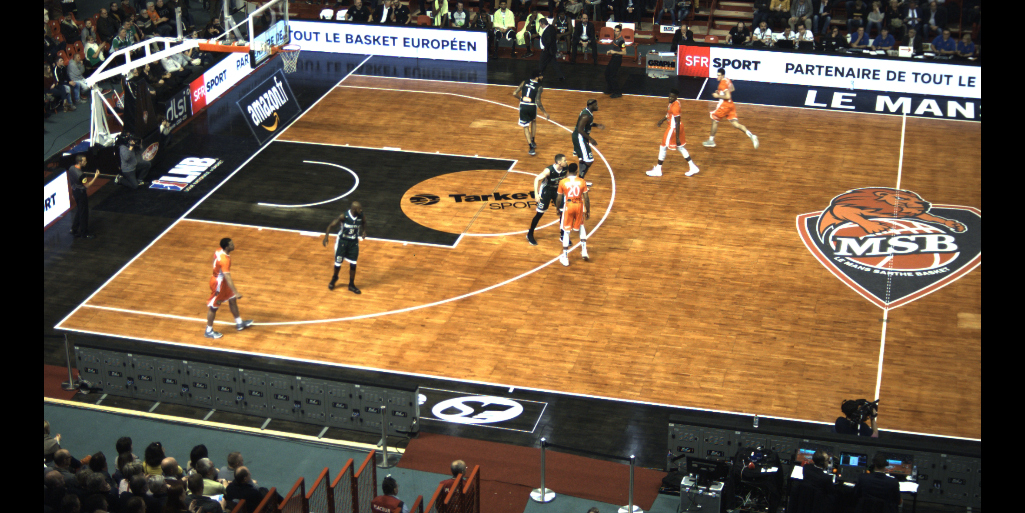} }
    \subfloat{ \includegraphics[width=0.3\linewidth,
                trim={65mm 60mm 95mm 20mm},clip]{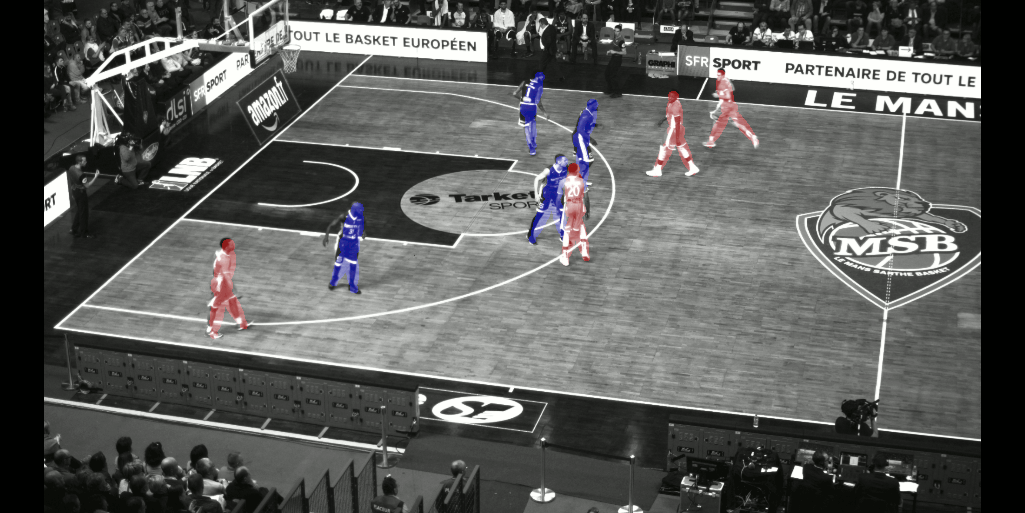} }
    \subfloat{ \includegraphics[width=0.3\linewidth,
                trim={65mm 60mm 95mm 20mm},clip]{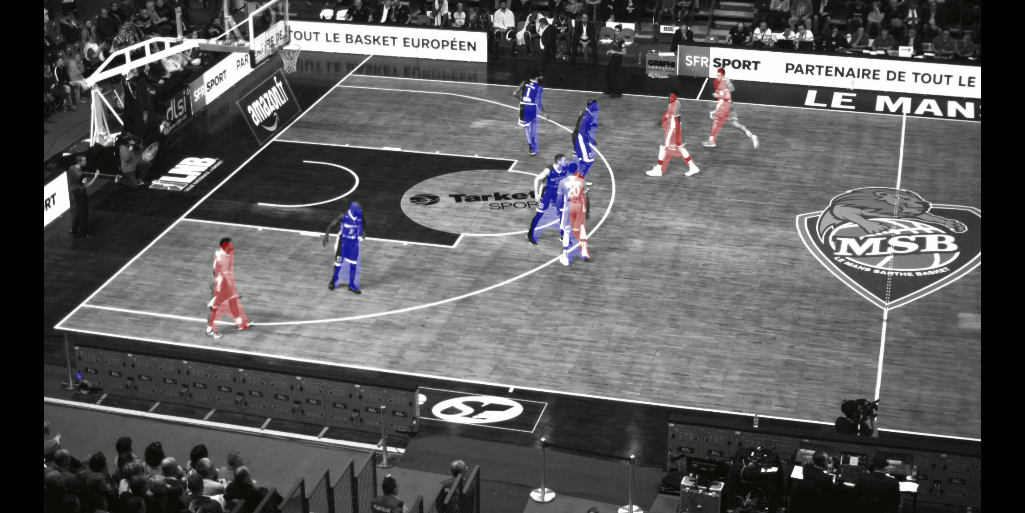} }
    \\
    \vspace{-0.3em}
    \subfloat{ \includegraphics[width=0.3\linewidth]{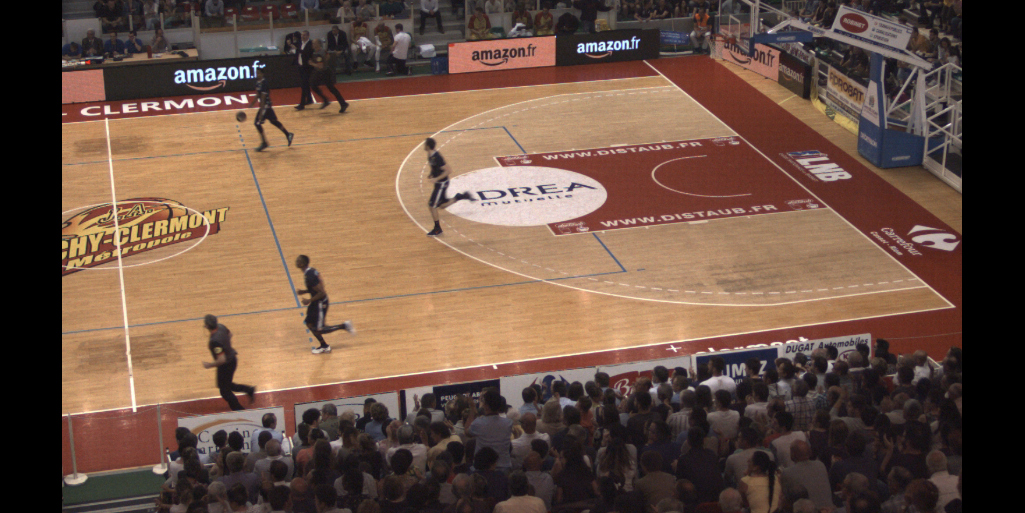} }
    \subfloat{ \includegraphics[width=0.3\linewidth,
                trim={20mm 35mm 80mm 15mm},clip]{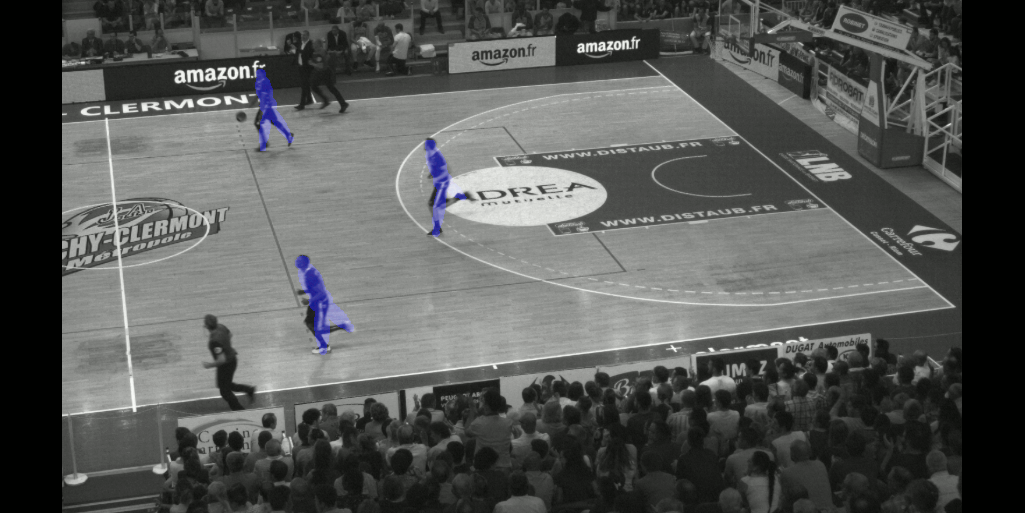} }
    \subfloat{ \includegraphics[width=0.3\linewidth,
                trim={20mm 35mm 80mm 15mm},clip]{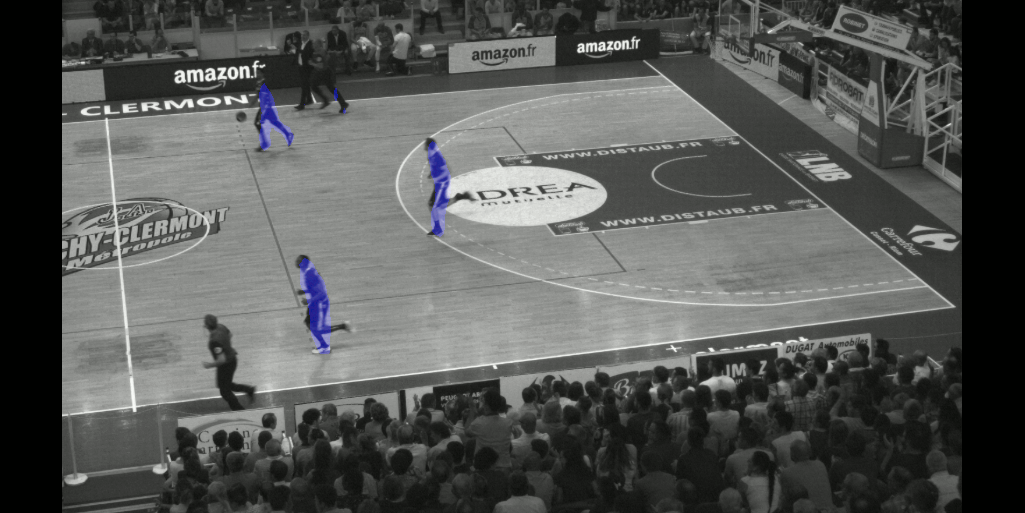} }
    \\
    \vspace{-0.3em}
    \noindent\rule{0.95\linewidth}{0.4pt}
    \subfloat{ \includegraphics[width=0.3\linewidth]{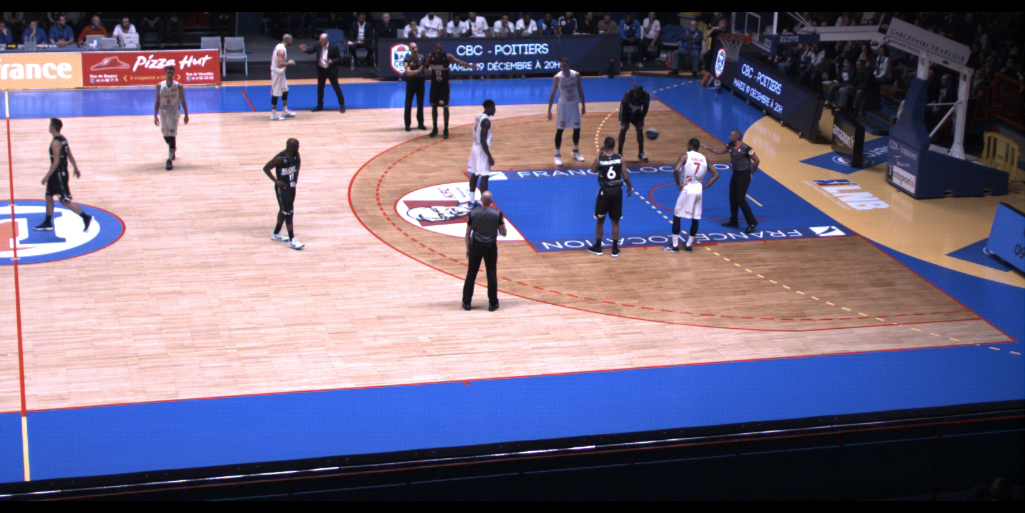} }
    \subfloat{ \includegraphics[width=0.3\linewidth,
                trim={10mm 50mm 90mm 0mm},clip]{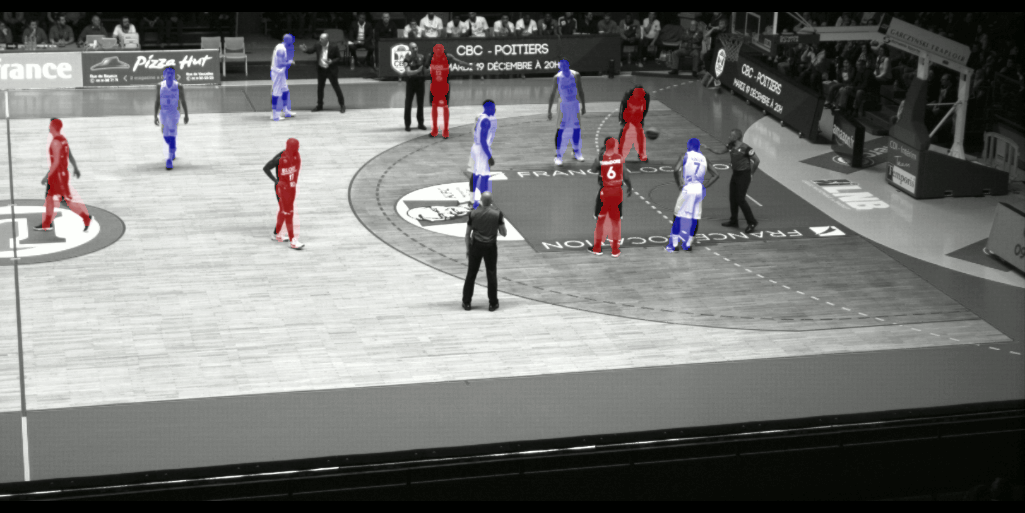} }
    \subfloat{ \includegraphics[width=0.3\linewidth,
                trim={10mm 50mm 90mm 0mm},clip]{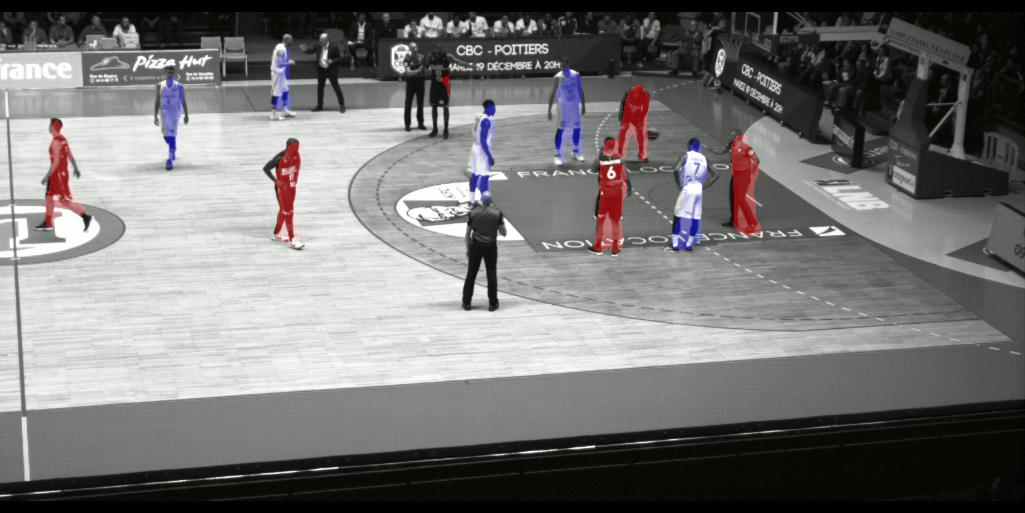} }
    \\
    \vspace{-0.3em} 
    \subfloat{ \includegraphics[width=0.3\linewidth]{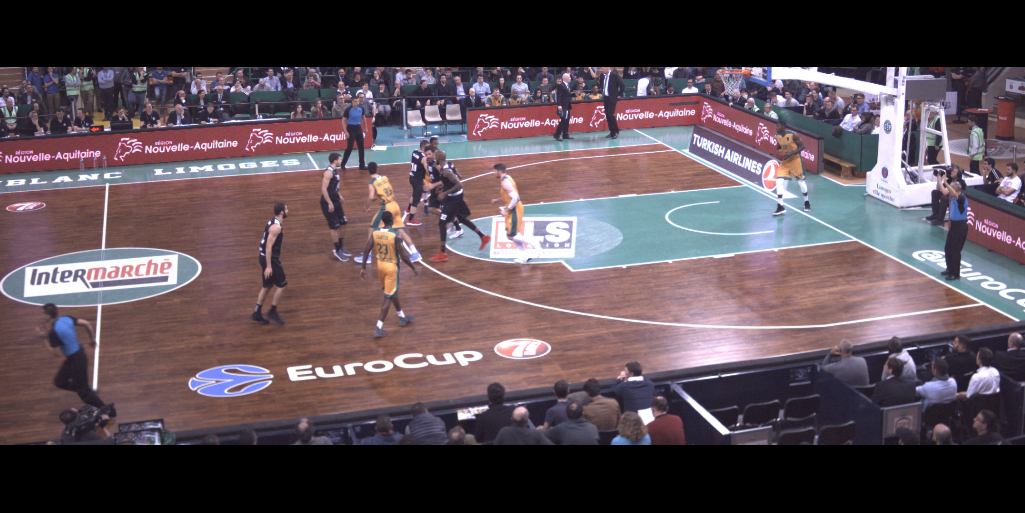} }
    \subfloat{ \includegraphics[width=0.3\linewidth,
                trim={10mm 50mm 140mm 25mm},clip]{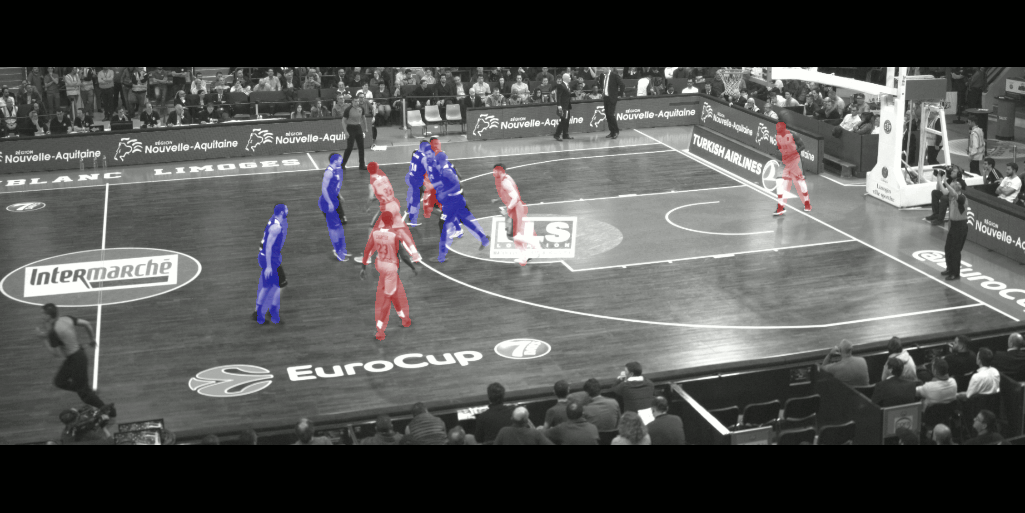} }
    \subfloat{ \includegraphics[width=0.3\linewidth,
                trim={10mm 50mm 140mm 25mm},clip]{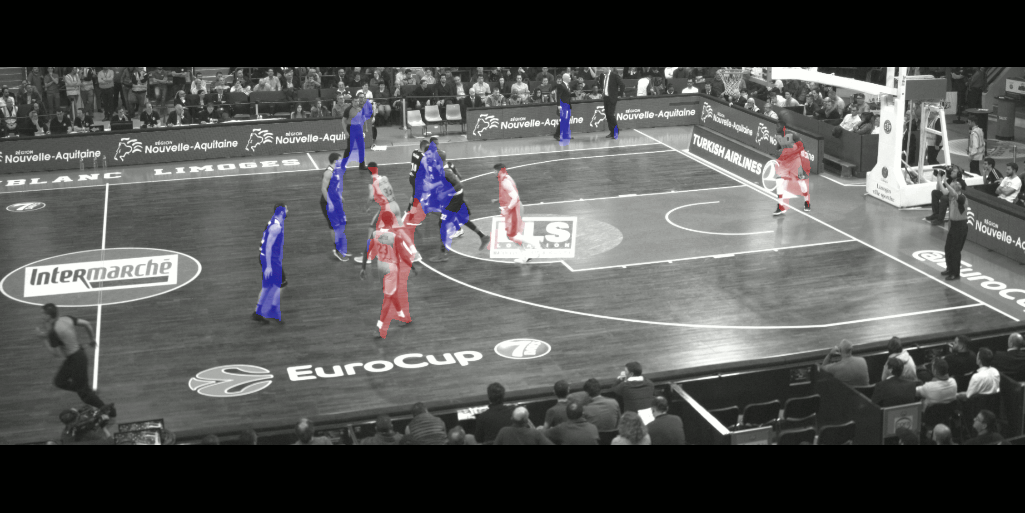} }
    \caption{Team discrimination with associative embedding. From left to right: test image, zoomed reference masks and prediction.
    The first five rows present success cases, while the last two show failure cases.
    From top to bottom:
    running players;
    strong shadows;
    occlusions;
    court and teams share the same colors;
    only one team; 
    confusion between players and referees;
    extreme occlusions. 
    Please refer to the numerical version of the paper for the colors and ability to zoom on details.
    }
    \label{fig:img-validation}
\end{figure*}

In Figure~\ref{fig:img-validation}, we observe that players are generally well detected but roughly segmented, probably due to our approximate training masks.
However, segmentation masks are very clean compared to the background-subtracted foreground masks derived for such kind of scenes (see for example \cite{Parisot17}). Therefore, they could advantageously replace those masks in algorithms using camera calibration to detect individual players from the segmentation mask~\cite{Alahi11,Carr12,Delannay09}.

In terms of team assignment, \cite{Lu18} mentions that they can not achieve good cross-game team assignment without fine-tuning.
In comparison, our method reaches more than 90\% of correct team assignments while testing on games and sport halls that are not seen during training.
The baseline Bayesian color histogram clustering only reaches 62\% of correct team assignments, which confirms that the team assignment task in the context of indoor sport is extremely difficult, as described in Section~\ref{sec:intro}.
We get near identical results for cross-arena evaluation.

\begin{table}
    \centering
    \begin{tabular}{|l|cc|}
        \hline
         Method & $R_{miss}$ & $R_{CTA}$ \\
         \hline
         \hline
         Associative Embedding & \multirow{2}{*}{0.11 $\pm$ 0.04} & 0.91 $\pm$ 0.04 \\
         Color Histogram &  & 0.62 $\pm$ 0.02 \\
         \hline
    \end{tabular}
    \caption{Evaluation measures on cross-game \mbox{K-fold}:
    mean and standard deviation of missed player detection and of correct team assignment rates, for 10 folds.}
    \label{tab:game-eval}
    \vspace{-0.3em}
\end{table}

\begin{table}
    \centering
    \begin{tabular}{|l|cc|}
        \hline
         Method & $R_{miss}$ & $R_{CTA}$ \\
         \hline
         \hline
         Associative Embedding & \multirow{2}{*}{0.11 $\pm$ 0.06} & 0.91 $\pm$ 0.03 \\
         Color Histogram &  & 0.63 $\pm$ 0.02 \\
         \hline
    \end{tabular}
    \caption{Evaluation measures on cross-arena \mbox{K-fold}:
    mean and standard deviation of missed player detection and of correct team assignment rates, for 10 folds.
    }
    \label{tab:hall-eval}
\end{table}

Qualitative results are shown in Figure~\ref{fig:img-validation}.
As written in Section~\ref{ssec:implem}, we intend to extract players only, excluding referees and other humans.
Images belong to testing folds, meaning that they originate from games or arenas not seen during training.
Teams masks are drawn in red and blue.

The first five rows in Figure~\ref{fig:img-validation} illustrate how well the proposed method can deal with indoor basketball conditions.
Players in fast movement and low contrast are detected and well grouped in teams.
Occlusions, led advertisements, and artificial lighting are not a major problem.
Associative embedding has a low sensitivity to high color similarities between background and foreground.
Specific treacherous scenes with players of only one team and some other humans are correctly handled.


We estimate to $10\%$ of the number of annotated players, the quantity of isolated regions that could fit humans, extracted in addition to reference instances.
These detections come from referees and other unwanted persons on or close to the ground, and in certain cases from scenery elements. 
In basketball, the proportion of the number of referees related to the players is from $20$ to $30\%$ (we usually count 2 or 3 referees in a complete field, while players are $5+5$).
Thus, it is interesting to see that our \mbox{FCN} trained on players generally avoids referees and other people.
However, this is a challenging task, as can be seen in the two prominent failure cases shown in the last two rows of Figure~\ref{fig:img-validation}, where referees shirts or pants are visually similar to a team.
In the first example, a referee is detected as a player and included in a team (referee on the right, under the basket),
and a player is filtered from predicted player class probably because it is seen as a referee by the network (background player in side of a referee).
In the second example, the dark pants of a referee and a coach in the back of the court are assimilated to the team in black.
This sample also presents a severe occlusion implying four players; inside and around this area, detection is inaccurate and team assignment of the orange player mixed with black teammates is lost.

\section{Conclusion}
\label{sec:conclu}

Associative embedding is considered to address the team assignment problem in team sport competitions.
It offers the advantage of discriminating teams in sport scenes, without requiring an unpractical per-game training.
Promising results are obtained on a challenging basketball dataset, with few tuning and only approximate player mask annotations.
In this work, the embeddings come with a player segmentation mask from a relatively simple multi-scale CNN, rather than the stacked hourglass network considered in previous works~\cite{Law18, Newell17b, Newell17a}.
Our work could be extended to support instance segmentation, by using either instance embeddings~\cite{Newell17a} or projective geometry~\cite{Alahi11,Carr12,Delannay09}.
Future investigations of interest include the explicit recognition of referees, a deeper analysis of the embeddings distribution and a more careful weighting of losses~\cite{Kendall18}.

{\small
\bibliographystyle{ieee}
\bibliography{biblio}
}

\end{document}